%% file: main.tex
\documentclass[10pt,twocolumn,letterpaper]{article}

\usepackage{iccv}      

\usepackage{algorithm}
\usepackage{algpseudocode} 
\usepackage{bm} 
\input{preamble}

\definecolor{iccvblue}{rgb}{0.21,0.49,0.74}
\usepackage[pagebackref,breaklinks,colorlinks,allcolors=iccvblue]{hyperref}

\def\paperID{11808} 
\def\confName{ICCV}
\def\confYear{2025}

\title{Analytic Subspace Routing: How Recursive Least Squares Works in Continual Learning of Large Language Model}

\author{\textbf{Kai Tong}$^{1}$, \textbf{Kang Pan}$^{2}$, \textbf{Xiao Zhang}$^{2}$, \textbf{Erli Meng}$^{2}$, \textbf{Run He}$^{1}$,  \textbf{Yawen Cui}$^{3}$, \\  \textbf{Nuoyan Guo}$^{1}$, \textbf{Huiping Zhuang}$^{1*}$ \\
$^{1}$ South China University of Technology \quad
$^{2}$ Xiaomi Corporation\\
$^{3}$ The Hong Kong Polytechnic University \\
{$^{*}$\tt\small corresponding: hpzhuang@scut.edu.cn}
}

\begin{document}
\maketitle
\input{sec/0_abstract}    
\input{sec/1_intro}

\input{sec/2_formatting}
\input{sec/3_finalcopy}

{
    \small
    \bibliographystyle{ieeenat_fullname}
    \bibliography{main}
}

\input{sec/X_suppl}

\end{document}

%% file: preamble.tex
\usepackage{tabularx}

%% file: sec/0_abstract.tex
\begin{abstract}

{Fine-tuning large language models (LLMs) on data of various domains encompassing advanced language-related tasks. Such a process is essentially a continual learning procedure since it is impractical to re-access the pre-trained data in most parties. This naturally attracts catastrophic forgetting of LLMs' previously learned knowledge, such as general skills. Existing techniques either leverage previous data to replay, leading to extra computational costs, or utilize a single parameter-efficient module to learn the downstream task, constraining new knowledge absorption with interference among different tasks. To handle these issues, we propose an Analytic Subspace Routing (Any-SSR) to address these challenges. For each task, we isolate the learning within a subspace of deep layers' features via low-rank adaptation, eliminating knowledge interference between different tasks. Additionally, we propose an analytic routing mechanism to properly utilize knowledge learned in different subspaces. Our approach employs Recursive Least Squares to train a multi-task router model, allowing the router to dynamically adapt to incoming data without requiring access to historical data. Subsequently, the router effectively assigns the current task to an appropriate subspace and has a non-forgetting property of previously learned tasks with a solid theoretical guarantee. Experimental results demonstrate that our method achieves near-perfect retention of prior knowledge while seamlessly integrating new information, effectively overcoming the core limitations of existing methods. Our code is available at 
\url{https://github.com/ZHUANGHP/Any-SSR}}
\end{abstract}

%% file: sec/1_intro.tex
\section{Introduction}
\label{sec:intro}

Recent years have witnessed remarkable advances in large language models (LLMs) \cite{gpt4,llama2,PaLM2024JMLR}, which have emerged as foundational technologies driving breakthroughs across diverse tasks, including natural language understanding \cite{llmunderstanding,LHPF}, multimodal reasoning \cite{fei-etal-2024-multimodal,multi-modal}, and application in robotics \cite{RAL2024,LLMRobotics2024ICRA}. However, a critical challenge persists in dynamic real-world environments, where new concepts and domains of knowledge emerge and vary frequently. To maintain LLMs' efficacy, models are required to absorb new domain knowledge through a fine-tuning (FT) process. Such a process essentially equals to continual learning (CL) of LLMs, with an aim to retain previously learned knowledge while accepting new-domain data.

A main unsolved problem in CL is the issue of \textit{catastrophic forgetting} \cite{cil_review2021NNs}, where the models could quickly lose the previous learned knowledge when accepting new tasks. This phenomenon is particularly severe in LLMs, since the general capabilities of LLMs encoded within high-dimensional parameter spaces can be easily disrupted by FT. This causes both prior task-specific knowledge and core generative competencies \cite{clllm}. Unlike traditional CL problems, prolonged iteration process in LLMs'
continual fine-tuning induces cumulative parameter drift. This could result in a severe decline of LLMs' task-specific abilities, as well as a significant portion of their general knowledge.


\begin{figure}[!h]
    \centering
    \includegraphics[width=\linewidth]{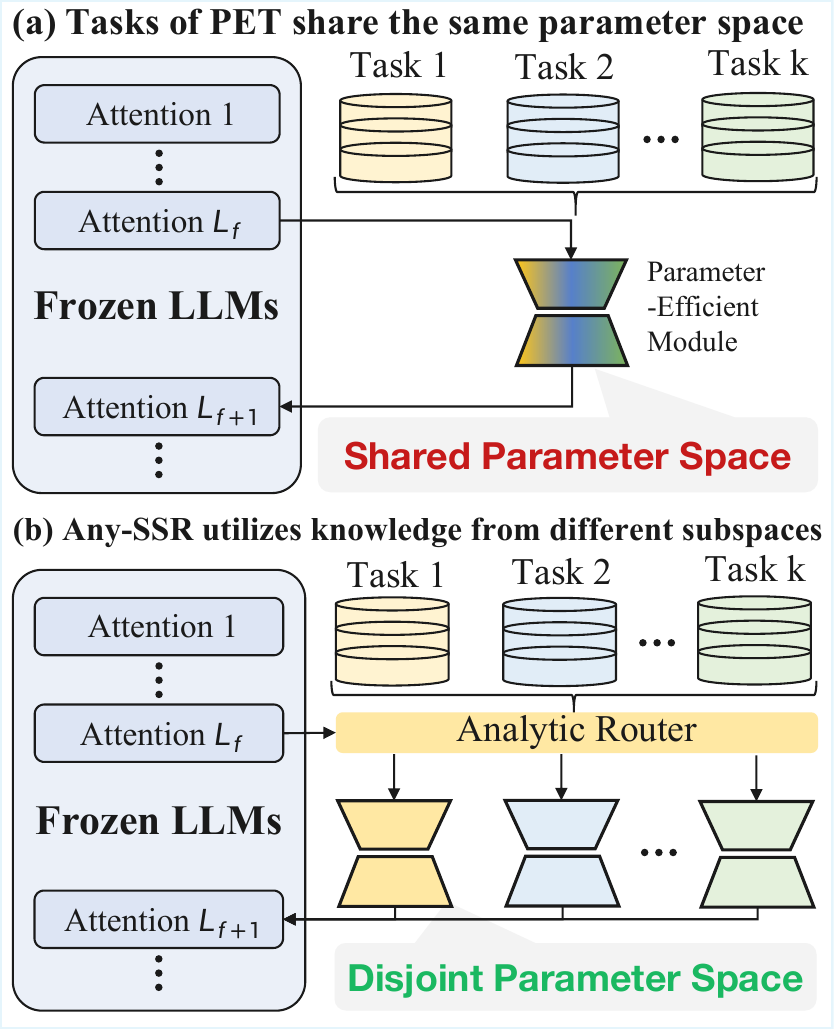}
    \caption{The difference between our method and parameter-efficient tuning (PET)-based approaches. While traditional fine-tuning introduces cumulative parameter drift that erodes both task-specific skills and core generative capabilities, our approach employs disjoint parameter spaces for individual tasks to avoid forgetting.}\label{fig:method-comparison}
\end{figure}

A traditional strategy to counter catastrophic forgetting is to store a small set of used data as exemplars for replaying \cite{he-etal-2024-seekr,leitner2024NAACL}. However, replaying past data for LLMs is less practical due to the huge volume of training data (especially those from pre-training). Moreover, replaying stored data could raise intensive computational cost and accompany potential privacy invasion. To avoid this, several techniques utilize parameter-efficient tuning (PET) \cite{PETSurvey2024TMLR} including prompt tuning and low-rank adaptation (LoRA) \cite{LoRA_Hu_ICLR2022} to continually adapt LLMs to new tasks. These methods mainly use light-weight module with a fixed number of parameters to absorb knowledge from all downstream tasks \cite{wang-etal-2023-orthogonal,qin2022lfpt5unifiedframeworklifelong}. Fine-tuning the shared module on one task followed by another consecutively inevitably induces catastrophic forgetting due to an extreme imbalanced data distribution \cite{FOAL2024NeurIPS}. It is hence highly motivated to explore new LLM technique that can continually absorb new skills in a replay-free fashion while overcoming the nature of catastrophic forgetting.

In this paper, we propose an Analytic Subspace Routing (Any-SSR) framework for continual learning of LLMs. Unlike common LLMs CL techniques, the proposed Any-SSR creatively adopts a classic recursive least squares (RLS) closed-form solution as an important mean to battle catastrophic forgetting. This comes in a hybrid structure, i.e., with an analytic router with a closed-form solution branching out in an earlier frozen segment of LLM layers, and the remaining layers tuned by LoRA using normal gradient-based technique. For each task, we isolate the learning within a subspace of deep layers' features via low-rank adaptation, eliminating knowledge interference among different tasks. We trained a specialized LoRA attachment, and use the analytic router to tell which one to load when inference. The key to avoid forgetting is that the analytic router bears the equivalence between continual training one task at a time, and the joint training by putting all-the-task data simultaneously. The main contributions of this work are listed as follow:
\begin{itemize}
        \item We introduce the Any-SSR, a novel closed-form subspace routing approach for continual learning in LLMs, facilitating dynamic and adaptable learning of new tasks.

        \item Our method structure each task as a subspace training using a specific LoRA model attachment, thereby reducing potential knowledge interference among tasks.
        
        \item Our method includes an analytic routing mechanism derived through recursive least squares, which seamlessly incorporate new tasks without forgetting. A theoretical guarantee of this non-forgetting property is provided to affirm its efficacy in mitigating catastrophic forgetting.
        
        \item We validate the Any-SSR through comprehensive experiments, achieving state-of-the-art performance on the Trace metric \cite{wang2023tracecomprehensivebenchmarkcontinual}.
        
    \end{itemize}

%% file: sec/2_formatting.tex
\section{Related Works}
\label{submission}
\subsection{Traditional Continual Learning}

Traditional continual learning aims at continuously updating models with data streams from scratch while addressing catastrophic forgetting during the continual learning process. Existing attempts on traditional continual learning can be divided into several categories. Replay-based methods \cite{rebuffi2017icarl, EEIL_2018_ECCV, LUCIR_Hou_CVPR2019, FOSTER2022ECCV, MRFA2024ICML} mitigate catastrophic forgetting by reintroducing part of past samples as exemplars to replay during training with several variants enhancing performance or optimizing memory usage. Regularization-based methods \cite{li2018LWF,EWC2017nas,CRNet2023TPAMI,RegCL2024ICML} impose additional constraints on network activations or key parameters to mitigate forgetting. Prototype-based methods \cite{Zhu_2021_CVPR,Petit_2023_WACV,PRAKA2023ICCV,FCS2024CVPR} employ prototypes to preserve decision boundaries across old and new tasks in the perspective of classifier training. Analytic learning-based methods \cite{zhuang2022acil, zhuang2023gkeal, Zhuang_DSAL_AAAI2024, FOAL2024NeurIPS} replace traditional backpropagation with a closed-form solution using techniques like Regularized Least Squares to achieve identical results of continual learning as those obtained by joint learning.  
However, implementing these techniques in LLMs poses several challenges. For instance, storing past data in replay-based methods can result in substantial computational expenses, determining parameter importance in regularization-based methods continues to be challenging due to the vast scale of LLMs, and prototype-based and analytic learning-based methods are oriented towards classifier learning, a component not present in LLMs. These methods also struggle with preserving general knowledge in LLMs post large-scale pretraining. To address these issues, we introduce Analytic Subspace Routing for continual learning in LLMs.

\subsection{Continual Learning of LLMs}
Due to the extensive scale of parameters in LLMs and the general skills acquired through intensive pre-training, CL in LLMs introduces additional challenges related to training costs and general knowledge retention. While various approaches have attempted to adapt traditional CL techniques to mitigate general knowledge forgetting, they are often hindered by the high training costs or the need to store previous data \cite{ILoRA, he-etal-2024-seekr}. For instance, SEEKR \cite{he-etal-2024-seekr} employs a method that involves the selection of attention heads based on forgettability and task-sensitivity metrics, followed by attention distillation on these heads using stored samples. Despite its ability to reduce the dependence on extensive data replaying, the complete fine-tuning process in SEEKR can still be costly. Various techniques have explored parameter-efficient tuning to alleviate training costs \cite{MoELoRA2024ACL, hu2021loralowrankadaptationlarge}. For example, O-LoRA focuses on LoRA-based architectures and facilitates continual learning within orthogonal subspaces \cite{hu2021loralowrankadaptationlarge}. Nevertheless, these methods typically rely on a fixed number of parameters to assimilate knowledge from subsequent tasks, resulting in ineffective adaptation to new tasks and potential interference between different tasks \cite{LBCL2024NeurIPS}. To tackle the challenges associated with high training costs and potential knowledge interference across tasks, we propose the Analytic Subspace Routing technique, which leverages parameter-efficient strategies and isolates subspaces to acquire distinct knowledge effectively. 

%% file: sec/3_finalcopy.tex
\section{The Proposed Method}

\subsection{Problem Definition}
The primary objective of CL for LLMs lies in optimizing their performance on sequentially acquired data streams while maintaining minimal degradation of their foundational competencies. 
Consider a sequence of new tasks denoted as $\{\mathcal{D}_{1}, \ldots,  \mathcal{D}_{k}\}$ arrived in a flow-like fashion. Each task $\mathcal{D}_{k} = \{(\bm{x}^i, \bm{y}^i)\}_{i = 1}^{n_k}$ contains $n_k$ pairs of question-answer($\bm{x}$, $\bm{y}$) data.
A single model needs to adapt to them sequentially with
only access to $\mathcal{D}_k$ at the $k$-th task. In general, given a prediction model $M_{\theta}$ parameterized by $\theta$, continual learning seeks to optimize for the following log-likelihood across all tasks:
\begin{equation}
\max_{\theta} \sum_{i = 1}^{k} \sum_{x,y \in \mathcal{D}_i} \log p_{\theta}(y | x).
\label{eq: optimizetarget}
\end{equation}

\begin{figure*}[!t]
    \centering
    \includegraphics[width=\linewidth]{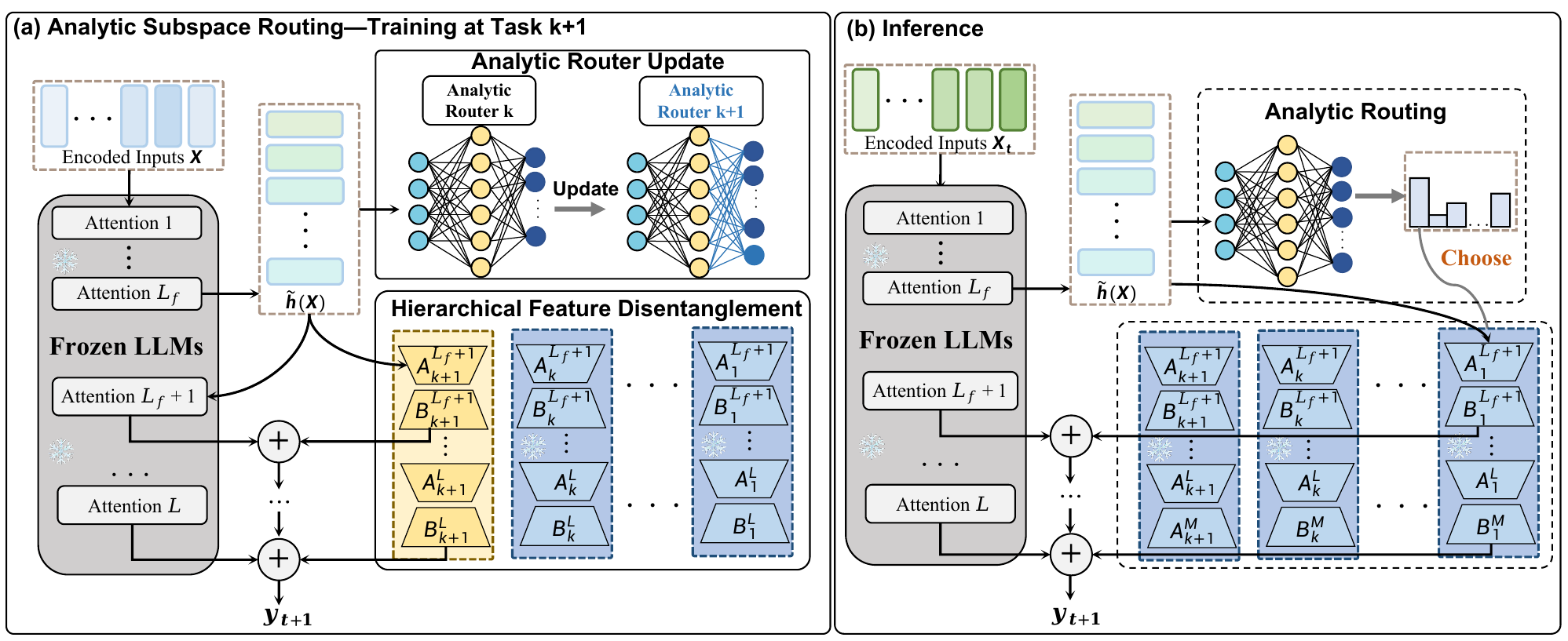}
    \caption{An overview of our proposed method Any-SSR. All the original parameters of the LLM are frozen. (a) The input data enters the first $L_f$ layers of the Transformer blocks in the model, and the feature $h_{\leq L_f}(\bm{x})$ are obtained. Features will be expanded to a larger dimension, then an analytic router will be trained in the manner of ridge regression. The trained router will perform recursive updates based on AutoCor $\bm{R}_k$ (or $\bm{R}_k$ and $\bm{Q}_k$) and the new coming dataset.(b) The inference module. The trained router accept the feature as input, and is responsible for selecting which LoRA task model to use during the inference process. }
    \vskip -0.1in
    \label{fig:mainflow}
\end{figure*}

Our approach incorporates the concept of recursion. In the following text, we first solve the scenarios with data $\{\mathcal{D}_{1}, \ldots,  \mathcal{D}_{k}\}$ using base training. Subsequently, we will illustrate how to recursively solve the problem when $\mathcal{D}_{k+1}$ arrives without leveraging the source data from $\mathcal{D}_1$ to $\mathcal{D}_k$.

\subsection{Overall Architecture}
We propose a closed-form LLM Continual Learning method, which achieves organic unification of shared representation and task-specific specialization through structural disentanglement. As shown in \ref{fig:mainflow}, the architecture comprises three core components: Frozen universal feature extractor, Task-specific Low-Rank Adaptation (LoRA) \cite{LoRA_Hu_ICLR2022} bank, and Recursive Analytic Learning(AL) based task router. For an input sequence the feedforward process is formalized as:

\begin{equation}
{y_{t+1}} = h_{\leq L_f}(\bm{x}
) \cdot f_{\theta_{k^*}}(h_{>L_f}(\bm{x})).
\label{main_process}
\end{equation}
Where $k^* = \arg\max_{k} g_k(h_{\leq L_f}(\bm{x}))$ indicates the selected task-id of the router, $h_{\leq L_f}$ denotes the frozen parameters of LLM's first $L_f$ layers, $h_{>L_f}$ represents the the base parameters of subsequent layers. $f_{\theta_k}$ is the LoRA adapter for task $k$, and $g_k$ denotes the normalized weights generated by the routing function. $\bm{x}$ is the input sequence of length $t$ and $y_{t+1}$ is the predicted next word. This design enables dynamic combination of base model capacity sharing and task-specific parameterization.



\subsection{Hierarchical Feature Disentanglement}

Inspired by hierarchical processing in cortical systems, we hypothesize that lower Transformer layers encode cross-task semantic features while higher layers handle task-specific semantic composition. Based on this, we partition parameters of the pre-trained LLM:

\begin{equation}
h_{\leq L_f} = \bigotimes_{l=1}^{L_f} \text{TransformerBlock}_l
\label{feature_layer}
\end{equation}

\begin{equation}
h_{>L_f} = \bigotimes_{l=L_f+1}^{L_{\text{total}}} \text{TransformerBlock}_l.
\label{train_layer}
\end{equation}
Here,  $h_{\leq L_f}(\bm{x}) \in \mathbb{R}^{t \times d}$ denotes the feature extracted by the low layers, $t$ denotes the length of the input $\bm{x}$ and $d$ is the hidden size of the transformer layers. $\bigotimes$ represents layer-wise composition, and  \text{TransformerBlock} indicates the frozen parameters of layer $l$. The benefit of disentanglement is that frozen lower layers preserve general linguistic understanding from pretraining.

For each new task $\mathcal{D}_k$, we insert a unique trainable LoRA adapter into shared higher layers. That is, given weight matrix $\bm{W}_l^{(k)}$ in layer $l$, LoRA decomposes parameter updates via low-rank projection:
\begin{equation}
\Delta \bm{W}_l^{(k)} = \bm{B}_l^{(k)}\bm{A}_l^{(k)},\quad \bm{B}_l^{(k)} \in \mathbb{R}^{d_{\text{in}}\times r},\ \bm{A}_l^{(k)} \in \mathbb{R}^{r\times d_{\text{out}}}
\label{lora}
\end{equation}
where $r \ll \min(d_{\text{in}}, d_{\text{out}})$ is the rank hyperparameter. The forward computation for a sequence $\bm{x}$ becomes:
\begin{equation}
f_{\theta_k}(h) = h_{>L_f}(\bm{x}) + \sum_{l=L_f+1}^{L_{\text{total}}} (\Delta \bm{W}_l^{(k)} h_l).
\label{lora_computation}
\end{equation}
Unlike conventional continual learning, all task adapters are trained independently with frozen base parameters to prevent interference.

\subsection{Analytic Routing Mechanism}
The router needs to processes low-layer features $h_{\leq L_f}(\bm{x})$ to predict task affiliation distribution $p(k|\bm{x})$, and update the weights without history data when new tasks are introduced. We propose a forgetting-free routing strategy based on ridge regression,
which consists of two core components: \textbf{Feature Expansion Module} that enhances feature separability and \textbf{Recursive Least-Square Solution} enabling incremental task learning.

Given frozen features $h_{\leq L_f}(\bm{X})$ (where $\bm{X}$ denotes all the samples from task 1 to $k$), we project them to an expanded space:
\begin{equation}
\tilde{h} = \phi(\text{mean-pool}(h_{\leq L_f}(\bm{X}))) \in \mathbb{R}^{E},
\label{feature_expand}
\end{equation}
where $\bm{X}$ is  in which mean-pool is used to Eliminate the dimension of the sequence length, that is:
\begin{equation}
\text{mean-pool}(\bm{H}) = \frac{1}{T} \sum_{t=1}^T \bm{H}_t,\quad \bm{H} \in \mathbb{R}^{T \times d}.
\label{mean-pool}
\end{equation}
Where $\phi: \mathbb{R}^d \to \mathbb{R}^E$ is a Gaussian-initialized, weight-frozen projection, implemented via linear transformation with ReLU activation($E >d$ as hyperparameter). The underlying cause is that expanding nonlinear feature expansion into higher-dimensional spaces exponentially improves linear separability probability, which is demonstrated by Cover's theorem \cite{cover}.

The calculation of the analytic router is a Linear Regression problem, the goal is to minimize:
\begin{equation}
\underset{\bm{W}_{\text{r}}}{\text{argmin}} (|| \bm{Y}_{1:k} - \tilde{h} _{1:k} \bm{W}^{\text{r}} ||_2^2 + \lambda || \bm{W}^{\text{r}} ||_2^2 ),
\label{goal_to_minimize}
\end{equation}
where $\lambda > 0$ is the coefficient of the regularization term, $\bm{W}^{\text{r}} \in \mathbb{R}^{E \times k}$ indicates the weight of the analytic router we aim to calculate and
\begin{equation}
            \tilde{h} _{1:k} = \begin{bmatrix}
                \tilde{h} _{1} \\
                \tilde{h} _{2} \\
                \vdots \\
                \tilde{h} _{k} \\
            \end{bmatrix},\qquad
            \bm{Y}_{1:k} = \begin{bmatrix}
                \bm{Y}_{1} \\
                \bm{Y}_{2} \\
                \vdots \\
                \bm{Y}_{k} \\
            \end{bmatrix}.
        \end{equation}
For which, $\bm{Y}_1, \dots, \bm{Y}_k$ are the task-IDs, which is the label of the router. Since the problem is a convex optimization problem, its solution can be obtained by equating its gradient to zero:
\begin{equation}
    \hat{\bm{W}}^{\mathrm{r}} = \left(\sum_{i=1}^{k}\tilde{h}_{i}^\mathrm{T} \tilde{h}_{i} + \lambda \bm{I}\right)^{-1} \left(\sum_{i=1}^{k}\tilde{h}_{i}^\mathrm{T} \bm{Y}_{i}\right)
    \label{solve}
\end{equation}
with $\bm{I}$ denotes the identity matrix. 

\subsection{Recursive Incremental Update}
Let $\bm{R}_k=\left(\sum_{i=1}^{k}\tilde{h}_{i}^\mathrm{T} \tilde{h}_{i} + \lambda \bm{I}\right)^{-1}$ and $\bm{Q}_k=\sum_{i=1}^{k}\tilde{h}_{i}^\mathrm{T} \bm{Y}_{i}$, where
$\bm{R}_k \in \mathbb{R}^{E \times E}$ is an autocorrelation matrix (AutoCor) and $\bm{Q}_k \in \mathbb{R}^{E \times k}$ is a cross correlation matrix (CrossCor).  $\bm{R}_k$ and $\bm{Q}_k$ capture the correlation information of $\tilde{h}_{1:k}$ and $\bm{Y}_{1:k}$, these matrix can recursively update the router weight ${\bm{W}}^\text{r}$ when new task $k+1$ arrives instead of using history datasets $\{\mathcal{D}_{1}, \ldots,  \mathcal{D}_{k}\}$.

Based on \eqref{solve}, we have:
\begin{equation}
    \hat{\bm{W}}^{\mathrm{r}}_{k+1} = \bm{R}_{k+1}\bm{Q}_{k+1}.
\end{equation}
According to the Woodbury matrix identity \cite{WoodburyIdentity_Woodbury1950}, for conformable matrices $\bm{A}$, $\bm{U}$, $\bm{C}$, and $\bm{V}$ , we have:
 \begin{equation}
                (\bm{A} + \bm{U}\bm{C}\bm{V})^{-1} = \bm{A}^{-1} - \bm{A}^{-1}\bm{U}(\bm{C}^{-1} + \bm{V}\bm{A}^{-1}\bm{U})^{-1}\bm{V}\bm{A}^{-1}.
            \end{equation}
Let $\bm{A} = \bm{R}_{k}^{-1}$, $\bm{U} = \tilde{h}_{k+1}^{\mathrm{T}}$, $\bm{V} = \tilde{h}_{k+1}$, and $\bm{C} = \bm{I}$, we can obtain that:
\begin{equation}
                \begin{split}
                    \bm{R}_{k+1} 
                    = \bm{R}_{k}    - \bm{R}_{k}\tilde{h} _{k+1}^{\mathrm{T}}(\bm{I} + \tilde{h}_{k+1}\bm{R}_{k}\tilde{h}_{k+1}^{\mathrm{T}})^{\mathrm{T}}\tilde{h}_{k+1}\bm{R}_{k}.
                \end{split}
                \label{rk+1}
            \end{equation}
According to the definition of $\bm{Q}_{k}$, the iterative expression of $\hat{\bm{W}}_{k+1}^{\text{R}}$ can be written as:
\begin{equation}
        \hat{\bm{W}}_{k+1}^{\text{R}} =  \bm{R}_{k+1}\bm{Q}_{k} + \bm{R}_{k+1} \tilde{h}_{k+1}^{\mathrm{T}} \bm{Y}_{k+1}.
                \label{wk+1}
            \end{equation}
Substitute Equ. \eqref{rk+1} into Equ. \eqref{wk+1}, we have:
\begin{align}\nonumber
\label{wk+1_2}
            \hat{\bm{W}}_{k+1}^{\text{r}}
            =& \bm{R}_{k}\bm{Q}_{k} + \bm{R}_{k} \tilde{h}_{k+1}^{\mathrm{T}} \bm{Y}_{k+1}  - \bm{R}_{k}\tilde{h}_{k+1}^\mathrm{T} (\bm{I} \nonumber \\ +& \tilde{h}_{k+1}\bm{R}_{k}\tilde{h}_{k+1}^{\mathrm{T}})^{-1}\tilde{h}_{k+1}\bm{R}_{k}\bm{Q}_{k} - \bm{R}_{k}\tilde{h}_{k+1}^\mathrm{T} (\bm{I} \nonumber \\ +& \tilde{h}_{k+1}\bm{R}_{k}\tilde{h}_{k+1}^{\mathrm{T}})^{-1}\tilde{h}_{k+1}\bm{R}_{k}\tilde{h}_{k+1}^T\bm{Y}_{k+1}.
            \end{align}
$\hat{\bm{W}}_{k+1}^{\text{R}}$ can be calculated using the previously AutoCor $\bm{R}_{k}$ and CrossCor $\bm{Q}_{k}$, as well as the features $\tilde{h}_{k+1}$ and labels $\bm{Y}_{k+1}$ of the newly incoming data. $\hat{\bm{W}}_{k+1}$ can also be expressed as:
\begin{equation}
                \hat{\bm{W}}_{k+1}^{\text{R}} = (\bm{I} - \bm{R}_{k+1}\bm{X}_{k+1}^{T} \bm{X}_{k+1})\hat{\bm{W}}_{k}^{\text{R}} + \bm{R}_{k+1}\bm{X}_{k+1}^{\mathrm{T}} \bm{Y}_{k+1}
                \label{eq:wk+1}
            \end{equation}
The proof can be found in the Supplementary material \ref{app:proof}.
            

By maintaining $\bm{R}_{k}$, our method achieves replay-free continual learning of the analytic router, which incrementally absorbs new tasks without forgetting historical knowledge.

\subsection{Inference Procedure}
The inference process enables task-aware generation through Analytic routing. Let input sequence be $\bm{X}_t = \{x_1,...,x_t\}$, which passes through frozen feature extractor:  
\begin{equation}  
   h_{\leq L_f}(X_t) = \bigotimes_{l=1}^{L_f} \text{TransformerBlock}_l(X_t),  
\end{equation}  
then followed by feature expansion:  
\begin{equation}  
   \tilde{h}(X_t) = \phi(\text{mean-pool}(h_{\leq L_f}(X_t))) \in \mathbb{R}^{d'}.  
\end{equation}
Compute task affiliation probability via analytic router weights:  
\begin{equation}  
   p(k|X_t) = \text{softmax}\left(\tilde{h}(X_t) W_r^{k}\right).
   \label{softmax}
\end{equation}  
Select dominant task expert:  
\begin{equation}  
     k^* = \arg\max_{k} p(k|X_t).  
     \label{selectdominant}
\end{equation}  
Activate corresponding LoRA adapter and get the next predicted token $y_{t+1}$:  
\begin{equation}  
     y_{t+1} \sim f_{\theta_{k^*}}(h_{>L_f}(X_t))  .
\end{equation}
Append generated token to the input sequence $X_t$ and repeat until $y_{t+1} = eos\_token$:  
\begin{equation}  
   X_{t+1} = X_t \oplus y_{t+1} .
\end{equation}
The proposed method preserves general knowledge via frozen base layers, achieves continual learning through analytic routing and incremental LoRA expert construction. The inference phase employs feature expansion and closed-form routing for task-aware generation. Our method have such advantages: (1) Forgetting prevention via parameter disentanglement and recursive updating; (2) Compact LoRA storage per task ($\mathcal{O}(rL_{\text{adapt}})$ parameters) reduces $>99\%$ storage vs full fine-tuning; (3) Historical data independence through AutoCor matrix storage; (4) CPU-compatible lightweight routing.

\begin{algorithm}[H]
\caption{The training process of the Analytic Router}\label{pseuducode}
\textbf{Analytic router training for ${\mathcal{D}_{1}, \ldots, \mathcal{D}_{k}}$}.
\begin{enumerate}
    \item Extract features $h_{\leq L_f}(\bm{X})$.
    \item Expand feature to $\tilde{h}(\bm{X})$ via \eqref{feature_expand}.
    
    \item \textbf{for} $i = 1$ to $k$ \textbf{do}:
        \begin{itemize}
            \item[] Train LoRA model $f_{\theta_i}(h)$ via \eqref{lora_computation}
        \end{itemize}
        \textbf{end for}
    \item Obtain analytic router weight $\hat{\bm{W}}^{\mathrm{r}}$ via \eqref{solve}.
    \item Save the AutoCor $\bm{R}_k$ and CrossCor $\bm{Q}_k$.
\end{enumerate}
\textbf{Continual Router Training for $\mathcal{D}_{k+1}$} 
\begin{enumerate}
    \item  Extract feature $h_{\leq L_f}(\bm{X}_{k+1})$.
    \item  Train LoRA model $f_{\theta_{k+1}}(h)$ via \eqref{lora_computation}
    \item  Obtain analytic router weight $\hat{\bm{W}}^{\mathrm{r}}$ via \eqref{solve}.
    \item  Save the AutoCor $\bm{R}_{k+1}$ and CrossCor $\bm{Q}_{k+1}$.
\end{enumerate}
\end{algorithm}

\begin{algorithm}[H]
\caption{Inference Procedure}\label{inference_pseudu} \label{loop}
\textbf{Inference}({$\bm{x}_t$}) \\
\textbf{while} {$y_{t+1} \neq \text{eos\_token}$} \textbf{do}
\begin{enumerate}
    \item Extract feature $h_{\leq L_f}(\bm{x}_{t})$ via $\bm{x}_t$
    \item Expand feature to $\tilde{h}(\bm{x}_{t})$ via \eqref{feature_expand}
    \item Calculate task probability via \eqref{softmax}
    \item Pick dominate task LoRA adapter via \eqref{selectdominant}
    \item Forward propagation: 
        $ y_{t+1} \gets f_{\theta_{k^*}}(h_{>L_f}(\bm{x}_t))$
    \item Update $\bm{x}_{t+1} \gets$ concat($\bm{x}_t, y_{t+1}$)
\end{enumerate}
\textbf{end while}
\end{algorithm}

\section{Experiments}
\subsection{Benchmarks}
We evaluate our proposed method using the TRACE benchmark \cite{wang2023tracecomprehensivebenchmarkcontinual}, a widely recognized framework for assessing continuous learning in LLMs. TRACE includes eight diverse datasets that span various domains, such as domain-specific knowledge, multilingual understanding, code generation, and mathematical reasoning. For our experiments, we use the reasoning-augmented versions of these datasets and follow the two task order configurations outlined in the original study. After the continuous learning process, we assess both the performance on the newly learned tasks and any changes in the general abilities of the LLMs. We follow SEEKER \cite{he-etal-2024-seekr} to employ two distinct training sequences:

\begin{table}[h]
\centering 
\caption{Two distinct CL task orders.}
\begin{tabularx}{\linewidth}{lX}
\toprule 
\textbf{\textcolor{blue}{Order1}}: & C-STANCE $\to$ FOMC $\to$ MeetingBank $\to$ Py150 $\to$ ScienceQA $\to$ NumGLUE-cm $\to$ NumGLUE-ds $\to$ 20Minuten \\
\midrule 
\textbf{\textcolor{red}{Order2}}: & NumGLUE-cm $\to$ NumGLUE-ds $\to$ FOMC $\to$ 20Minuten $\to$ C-STANCE $\to$ Py150 $\to$ MeetingBank $\to$ ScienceQA \\
\bottomrule 
\end{tabularx}
\label{tab:task_orders}
\end{table}

\subsection{Baselines and Evaluation Metrics}
In this paper, we employ both the overall performance (OP)\cite{Chaudhry_2018_ECCV} and the backward transfer scores (BWT)\cite{NIPS2017_f8752278} as the principal evaluation metrics. For tasks $\{\mathcal{D}_{1}, \ldots,  \mathcal{D}_{k}\}$, upon incrementally learning the m-th task $\mathcal{D}_k$, the score achieved by the model on the i-th task $\mathcal{D}_i$ is represented as $A_{i,t}$ (where $t \geq i$). The computations for the overall performance and the backward transfer score are carried out using the subsequent formulas:

\begin{equation}
OP = \frac{1}{k} \sum_{i = 1}^{k} A_{i,k} \quad 
\end{equation}

\begin{equation}
BWT = \frac{1}{k - 1} \sum_{i = 1}^{k - 1} (A_{i,k} - A_{i,i}) \quad 
\end{equation}

We compare our proposed method, Any-SSR, with ten baseline approaches commonly used in the context of continuous learning:

\textbf{SeqFT}: Sequential fine-tuning of the model without employing any specific strategies for continuous learning.
\textbf{EWC} \cite{EWC2017nas}: Regularizes parameter updates by constraining changes based on their importance scores, mitigating catastrophic forgetting.
\textbf{LwF} \cite{li2018LWF}: Uses knowledge distillation, where the model for the previous task is distilled into the current task using the current task’s data.
Re-play: Fine-tunes the model on the current task’s data along with a small set of replay samples from previous tasks.
\textbf{DER++} \cite{NEURIPS2020_b704ea2c}: Stores model predictions (logs) for replay samples from previous tasks and combines this with knowledge distillation and replay to reduce forgetting.
\textbf{LFPT5}\cite{qin2022lfpt5unifiedframeworklifelong}: Learns a soft prompt to generate pseudo-samples of previous tasks, enabling replay without explicitly storing past data.
\textbf{O-LoRA} \cite{wang-etal-2023-orthogonal}: Applies orthogonal constraints to LoRA matrices for each task, ensuring task-specific updates while reducing interference.
\textbf{L2P} \cite{Wang_2022_CVPR}: Maintains a prompt pool to dynamically select and fine-tune prompts specific to individual samples.
\textbf{PP} \cite{razdaibiedina2023progressivepromptscontinuallearning}: Tunes a set of prompts for each task and concatenates them during inference to encode task-specific knowledge.
\textbf{SEEKR} \cite{he-etal-2024-seekr}: conducts attention distillation on the selected attention heads to recognize the most valuable attention heads.We include the results of \textbf{multi-task learning} (MTL) as an upper-bound reference, representing the ideal scenario where the model is trained on all tasks simultaneously.
\cite{AANet_2021_CVPR}

\subsection{Training details}
Our experiments were conducted on the Llama2-7B-chat foundation model\cite{llama2} using the HuggingFace Transformers implementation. Following Trace\cite{wang2023tracecomprehensivebenchmarkcontinual}, We maintained a consistent training sample size of 5,000 with a constant learning rate of 1e-5 across all tasks. The model was trained on eight benchmark datasets spanning multiple domains: C-STANCE(social media stance detection), FOMC (financial decision parsing), MeetingBank (meeting summarization), Py150 (code generation), ScienceQA (science question answering), and two numerical reasoning datasets (NumGLUE-cm, NumGLUE-ds), along with the 20Minuten news comprehension corpus. Since our approach can overcome forgetting, we trained the model for 10 epochs compared to Trace's 5 epochs.

The optimization configuration employed AdamW with weight decay disabled and a batch size of 128 sequences. All experiments were executed on a compute cluster equipped with 7×NVIDIA GeForce RTX 4090 GPUs (24GB VRAM each) using DeepSpeed ZeRO-3 optimization.

\subsection{Experimental Result Analysis}

\begin{table}[t]
\centering
\caption{Experiments on the TRACE benchmark using LLaMA-2-7B-Chat. The results in \% are generated by two different transfer orders, and are displayed in the format of OP (BWT).}
\footnotesize
\captionsetup{font=footnotesize}
\begin{tabular}{lcccc}
\toprule
& Order1 & Order2 \\
\midrule
SeqFT & $47.63\ (-11.45)$ & $45.12\ (-12.27)$ \\
EWC & $48.20\ (-9.48)$ & $44.54\ (-12.00)$ \\
LwF & $41.86\ (-6.50)$ & $40.25\ (-5.96)$ \\
LFPT5 & $38.67\ (-11.43)$ & $42.26\ (-7.43)$ \\
L2P & $35.23\ (-15.96)$ & $34.63\ (-16.86)$ \\
PP & $29.41\ (-5.79)$ & $21.58\ (-8.83)$ \\
O-LoRA & $44.64\ (-4.20)$ & $42.83\ (-9.11)$ \\
\textbf{Any-SSR} & $\mathbf{55.69}\ (-0.00)$ & $\mathbf{55.69}\ (-0.00)$ \\
\multicolumn{3}{c}{Exemplar-free methods} \\ 
\midrule
Replay (1\%) & $48.47\ (-9.69)$ & $47.04\ (-10.24)$ \\
DER++ (1\%) & $49.22\ (-8.32)$ & $46.59\ (-10.91)$ \\
SEEKR (1\%) & $54.99\ (-2.61)$ & $54.69\ (-2.53)$ \\
\multicolumn{3}{c}{Replay-based methods} \\ 
\midrule
Upper-bound & \multicolumn{2}{c}{59.38} \\
\bottomrule
\end{tabular}
\label{tab:baseline}
\end{table}

\subsubsection{Performance on Continual Learning Tasks}
To validate the performance of continual learning, we compared the proposed Any-SSR with baseline methods on the TRACE benchmark using two distinct learning orders. Table \ref{tab:baseline} illustrates that Any-SSR demonstrates competitive performance in both OP and BWT with significant margins when compared to exemplar-based methods. For instance, when assessed with Order1, Any-SSR achieves an OP of 55.69\%, surpassing the second exemplar-free method, SepFT, by 8.06\%. This trend persists when evaluating with Order2. When comparing with replay-based methods, although replay-based methods take advantage of using exemplars and achieve general good results when compared with exemplar-free methods except Any-SSR, Any-SSR can still outperform the state-of-the-art repaly-based methods. That is, Any-SSR achieves superior result compared with the state-of-the-art SEEKER with both Order1 and Order2 (55.69\% and 55.69\% v.s. 54.99\% and 54.69\%). This leading pattern demonstrates the competitiveness of Any-SSR in adapting LLMs to downstream tasks. Furthermore, Any-SSR stands out as the sole method achieving 0 BWT. This observation showcases Any-SSR's ability to prevent interference between tasks emphasizing its superior knowledge retention. Notably, Any-SSR consistently produces identical results when learning with different orders, while other compared methods are susceptible the order of learning tasks. This observation demonstrates Any-SSR's robustness and its invariance to task order.

\begin{figure*}[h!]  
    \centering  
    \includegraphics[width=\linewidth]{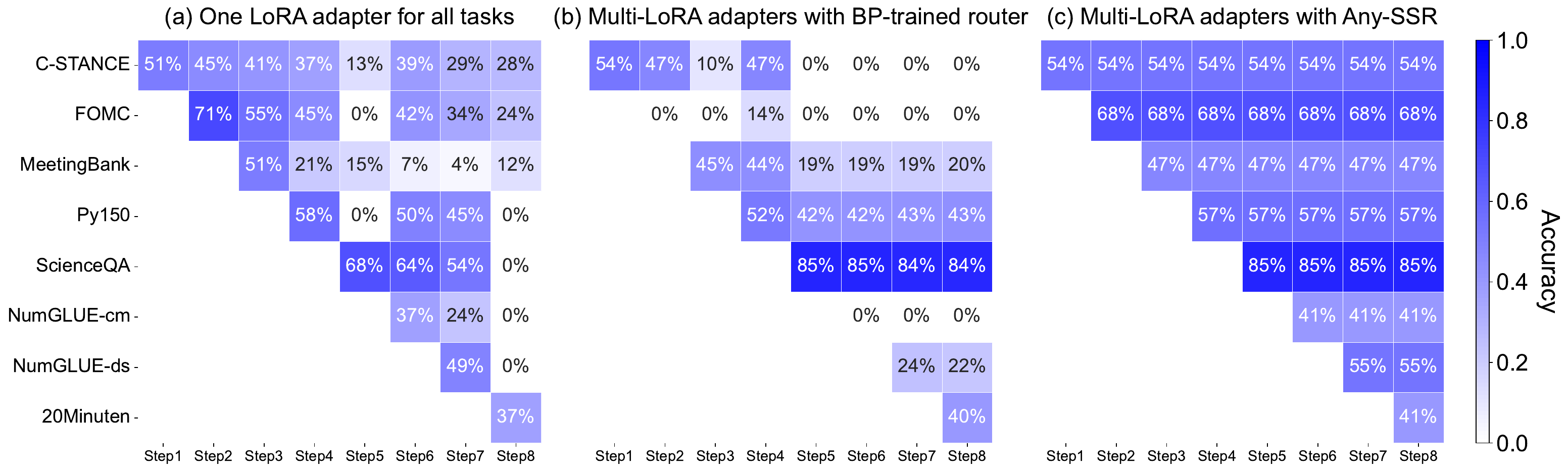}
    \caption{Heatmap comparing the accuracy of different methods across sequential tasks. Deep colors (blue) indicate higher accuracy rates, while light colors (white) represent lower performance.}
    \label{fig:heatmap}  
\end{figure*} 

\subsubsection{Ablation study }
\textbf{Ablation on Components in Any-SSR.} To provide a systematical analysis on each component in Any-SSR, we compare our method with two other strategies to demonstrate the effect of the isolated subspace training and analytic router. The alternative methodsinvolve employing a single LoRA adapter for multi-task continual learning and utilizing a distinct LoRA adapter for each task while training the router using BP. For LoRA training, we set the rank as 8 while the learning rate as 1e-4; for BP-based router training, we utilize a two-layer MLP with the same expansion size $E$ as AL-based router. As shown in Figure \ref{fig:heatmap}, our method demonstrates performance advantage over directly using a single LoRA and training a router with BP. For the methods of single LoRA, at each step, the new task can obtain competitive results while previous tasks encounter performance drop. This observation indicates the knowledge interference across different tasks. For the BP-based router, results display a more erratic pattern. To delve deeper into this phenomenon, we conducted additional experiments on these two routers.

\begin{figure}[h]
    \centering
    \includegraphics[width=\linewidth]{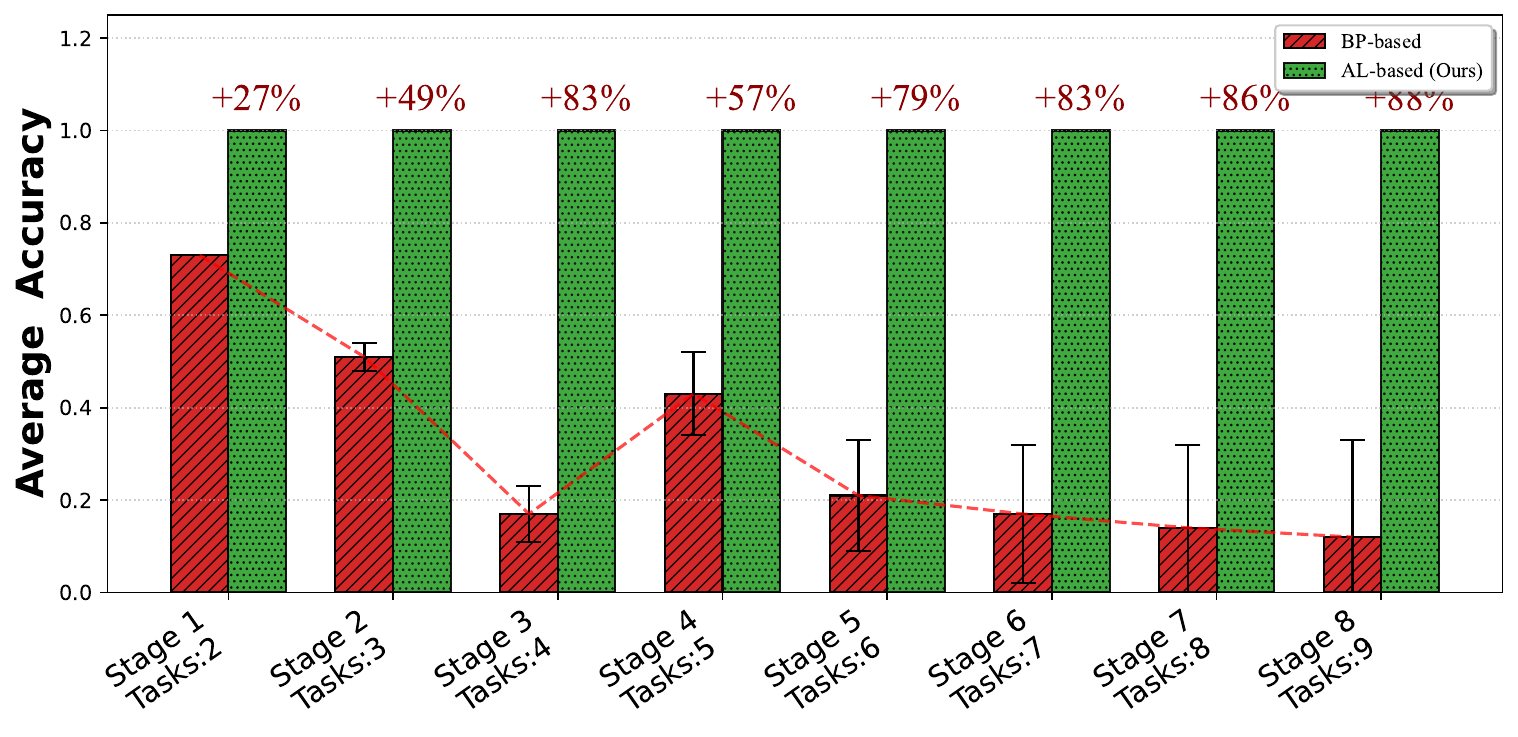}
    \caption{Comparison of the router's Acc in \%. using Analytic Training and directly BP training. The horizontal axis denotes 8 incremental training stages (introduce a new task in each stage), and the vertical axis shows the average accuracy of all current tasks} 
    \vskip -0.1in
    \label{fig:routeracc}
\end{figure}

\textbf{AL Contributes to a Perfect Routing.} To visualize the routing results of routers obtained with different methods, we report the routing accuracy at each step in Figure \ref{fig:routeracc}. For the BP-based router, it suffers progressive performance degradation (21.7\% average accuracy drop per phase). Subsequently during the inference, the BP-based router can assign wrong subspaces to current task, leading to disrupted answer patterns for certain tasks and causing the performance of these tasks to drop to zero. In contrast, the analytic router showcases zero-forgetting attributes throughout sequential training phases, maintaining 100\% accuracy on previous tasks throughout the continual learning process. This stark difference highlights that the performance deterioration of the BP router stems from catastrophic forgetting. Our method benefits from the non-forgetting routers and achieves competitive results on all tasks.

\textbf{Analysis on Hyperparameters in Any-SSR.} As shown in \ref{tab:LE}, we present the performance of Any-SSR with different feature layer $L_f$ and scale-up dimension $E$. Regarding $L_f$, lower feature layers (e.g., $L_f=2$) exhibit non-zero BWT and lower OP, possibly due to the general nature of these lower features that compromising the router's accuracy. Performance initially improves with an increase in $L_f$, but starts to degrade beyond $L_f > 4$. This decline can be attributed to the decreased effectiveness of the remaining layers in task-specific learning with deeper feature layers, leading to a decrease in OP.Regarding $E$, it is noticeable that performance increases as $E$ scales up from 5k to 10k when $L_f=4$. However, further increases in $E$ do not result in performance enhancements, indicating that choosing $E=10k$ could be sufficient. The configuration of $L_f = 4, E = 10k$ dimensions yields optimal results.



\begin{table}[h]
\caption{The performance (OP(BWT)) using different layers $L_f$ and Feature Expanding size $E$ on TRACE order1.}
\centering
\footnotesize
\captionsetup{font=footnotesize}
\setlength{\tabcolsep}{8pt}
\begin{tabular}{lcccc}
\toprule
$L_f$ & $E$ & OP(BWT) \\
\midrule
2 & 5000 & 54.76(-0.21) \\
2 & 10000 & 54.79(-0.19) \\
2 & 15000 & 54.79(-0.19) \\
\hline
4 & 5000 &  54.89(-0.12)\\
\textbf{4} & \textbf{10000} & \textbf{55.69(0.00)} \\
4 & 15000 & 55.69(0.00) \\
\hline
6 & 5000 &  53.21(0.00)\\
6 & 10000 &  53.21(0.00)\\
6 & 15000 &  53.21(0.00)\\
\bottomrule
\end{tabular}

\label{tab:LE}
\end{table}

\subsubsection{General Ability after Continual Learning}

To validate Any-SSR's performance of preserving general knowledge after continual learning in downstream tasks, we report the performance of models after CL on general skills benchmarks. It is crucial to emphasize that our Any-SSR utilizes only 1\% of the merged validation subset data across all evaluated tasks exclusively for training the router module. This limited dataset is strictly employed to optimize the router’s decision boundaries through constrained gradient updates, directing the general questions to the origin model with parameters entirely isolated from any exposure to these samples.

As shown in Table \ref{generalability}, Any-SSR demonstrates relatively stable performance across various tasks. When compared with exemplar-free method SeqFT, Any-SSR can maintain better general skill. Although the replay-based methods (i.e., Replay and SEEKER) outperforms Any-SSR, the gaps are relative small. This observation demonstrates the effectiveness of Any-SSR to retain general knowledge while adhere to exemplar-free constrain. This phenomenon stems from two methodological characteristics:



\begin{table}[!h]
\centering
\footnotesize
\captionsetup{font=footnotesize}
\caption{General abilities after continual learning with different methods with performance averaged over two task orders}
\label{generalability}
\resizebox{0.48\textwidth}{!}{
\begin{tabular}{cccccccc}
\toprule
& \textbf{MMLU} & \textbf{GSM} & \textbf{BBH} & \textbf{TydiQA} & \textbf{BoolQ} & \textbf{PIQA} & \textbf{GA (DeltaGA)} \\
\hline
LLaMA-2-7B-Chat & 46.89 & 27.14 & 39.73 & 16.76 & 79.79 & 76.33 & 47.77 \\

SeqFT & 45.16 & 14.03 & 32.50 & 14.84 & 79.00 & 75.49 & 43.50 (-4.27) \\

Replay (1\%) & 45.49 & 12.70 & 33.46 & 14.65 & 78.69 & 75.65 & 43.44 (-4.33) \\

SEEKR (1\%) & 46.32 & 20.85 & 38.52 & 18.22 & 80.64 & 75.79 & 46.72 (-1.05) \\
\hline



\textbf{Any-SSR} & 45.77 & 25.43 & 37.01 & 15.52 & 79.67 & 75.67 & 46.51 (-1.26) \\
\bottomrule
\end{tabular}
}
\end{table}

\section{Conclusion}
In this work, we propose Any-SSR, a novel framework that enables efficient, memory-free, and scalable continual learning for LLMs. By leveraging Recursive Least Squares (RLS) for a dynamic multi-task routing mechanism, Any-SSR eliminates the need for storing or replaying historical data, addressing core limitations of traditional methods. Through a combination of parameter-efficient adaptation using LoRA and an analytic routing approach, Any-SSR achieves near-perfect retention of prior knowledge while seamlessly integrating new tasks, demonstrating robust and efficient performance. Our approach not only overcomes the core limitations of existing methods such as catastrophic forgetting and computational inefficiency, but also provides a solid theoretical foundation for the non-forgetting property, ensuring stability and reliability in continual learning scenarios.

%% file: sec/X_suppl.tex
\clearpage
\setcounter{page}{1}
\maketitlesupplementary
\appendix
\renewcommand{\thesection}{\Alph{section}}
\section{Proof of Eqa. \eqref{eq:wk+1}}\label{app:proof}
Proof. According to \ref{wk+1}:

\begin{equation}\label{wk}
    \hat{\bm{W}}_{k+1}^{\text{r}} = \bm{R}_{k+1}\bm{Q}_{k+1} = \bm{R}_{k+1}\bm{Q}_{k} + \bm{R}_{k+1} \tilde{h}_{k+1}^{\mathrm{T}} \bm{Y}_{k+1}.
\end{equation}
Noticed that $\bm{R}_{k+1}\bm{Q}_{k+1}$ can be written as:
\begin{align*}
\bm{R}_{k+1}\boldsymbol{Q}_{k}&=\boldsymbol{R}_{k}\boldsymbol{Q}_{k}-\boldsymbol{R}_{k}\tilde{h}_{k+1}^{\top}\boldsymbol{K}\tilde{h}_{k+1}\boldsymbol{R}_{k}\boldsymbol{Q}_{k}\\
&=(\boldsymbol{I}-\boldsymbol{R}_{k}\tilde{h}_{k+1}^{\top}\boldsymbol{K}_{k+1}\tilde{h}_{k+1})\hat{\boldsymbol{W}}_{k},
\end{align*}

Where 
$\boldsymbol{K}=(\boldsymbol{I}+\tilde{h}_{k+1}\boldsymbol{R}_{k}\tilde{h}_{k}^{T})^{-1}$ 
and 
$\boldsymbol{K}\in\mathbb{R}^{n \times n}$.

Since
\begin{equation}
\boldsymbol{K}\boldsymbol{K}^{-1}=\boldsymbol{K}(\boldsymbol{I}+\tilde{h}_{k+1}\boldsymbol{R}_{k}\tilde{h}_{k+1}^T)=\boldsymbol{I},
\end{equation}
we have 
\begin{equation}
   \boldsymbol{K}=\boldsymbol{I}-\boldsymbol{K}\tilde{h}_{k+1}\boldsymbol{R}_{k}\tilde{h}_{k+1}^{\top}.
\end{equation}

Therefore

\begin{equation}
    \boldsymbol{R}_{k+1}\boldsymbol{Q}_{k}=(\boldsymbol{I}-\boldsymbol{R}_{k+1}\tilde{h}_{k+1}^T\tilde{h}_{k+1})\hat{\bm{W}}_{k}^{\text{R}}
    \label{21}.
\end{equation}

Instantiate \ref{21} into \ref{wk+1}, we have 

\begin{equation}
                \hat{\bm{W}}_{k+1}^{\text{r}} = (\bm{I} - \bm{R}_{k+1}\tilde{h}_{k+1}^{T} \tilde{h}_{k+1})\hat{\bm{W}}_{k}^{\text{R}} + \bm{R}_{k+1}\tilde{h}_{k+1}^{\mathrm{T}} \bm{Y}_{k+1}.
            \end{equation}

which completes the proof.